\documentclass{article}
\pdfoutput=1
\usepackage[T1]{fontenc}
\usepackage[utf8]{inputenc}
\usepackage{float}
\usepackage{url}
\usepackage{amsmath}
\usepackage{graphicx, wrapfig}
\usepackage{microtype}
\usepackage[unicode=true,
 bookmarks=false,
 breaklinks=false,pdfborder={0 0 1},backref=false,colorlinks=false]
 {hyperref}

\makeatletter

\floatstyle{ruled}
\newfloat{algorithm}{tbp}{loa}
\providecommand{\algorithmname}{Algorithm}
\floatname{algorithm}{\protect\algorithmname}

     \PassOptionsToPackage{numbers, compress}{natbib}

     \usepackage[final]{ml4eng_2020}

\usepackage{url}
\usepackage{booktabs}
\usepackage{amsfonts}
\usepackage{nicefrac}

\title{Analog Circuit Design with Dyna-Style Reinforcement Learning}

\author{
  Wook Lee \\
  Department of Intelligent Systems \\
  Delft University of Technology \\
  \texttt{wleegt@gmail.com} \\
  \And
  Frans A. Oliehoek \\
  Department of Intelligent Systems \\
  Delft University of Technology \\
  \texttt{f.a.oliehoek@tudelft.nl} \\
}

\makeatother

\begin{document}
\maketitle
\begin{abstract}
In this work, we present a learning based approach to analog circuit
design, where the goal is to optimize circuit performance subject
to certain design constraints. One of the aspects that makes this
problem challenging to optimize, is that measuring the performance
of candidate configurations with simulation can be computationally
expensive, particularly in the post-layout design. Additionally, the
large number of design constraints and the interaction between the
relevant quantities makes the problem complex. Therefore, to better
facilitate supporting the human designers, it is desirable to gain
knowledge about the whole space of feasible solutions. In order to
tackle these challenges, we take inspiration from model-based reinforcement
learning and propose a method with two key properties. First, it learns
a reward model, i.e., surrogate model of the performance approximated
by neural networks, to reduce the required number of simulation. Second,
it uses a stochastic policy generator to explore the diverse solution
space satisfying constraints. Together we combine these in a Dyna-style
optimization framework, which we call DynaOpt, and empirically evaluate
the performance on a circuit benchmark of a two-stage operational
amplifier. The results show that, compared to the model-free method
applied with 20,000 circuit simulations to train the policy, DynaOpt
achieves even much better performance by learning from scratch with
only 500 simulations.
\end{abstract}

\section{Introduction}

\label{introduction}

Although analog circuits are present as many different functional
blocks in most integrated circuit (IC) chips nowadays, analog circuit
design has been increasingly difficult for several reasons, e.g.,
growing complexity in circuit topology, tight tradeoffs between different
performance metrics and so on \cite{allen2002cmos}. Unlike digital
circuit design that is aided by well standardized design flow with
electronic design automation (EDA) tools \cite{rabaey2003digital},
analog design requires a high level of application-specific customization
and still resorts to domain knowledge by experienced designers. Manual
optimization based on exhaustive search with simulation is a notoriously
time consuming and labor intensive task. Particularly the post-layout
design, in which the design objectives are verified after the layout
creation, is much more challenging because it demands additional computationally
costly simulations using EDA tools to reflect the layout parasitics
to the circuit accurately. In addition, the circuit problem can have
multiple different solutions which are distributed diversely in the
search space. It can serve as crucial information fed back to designers
such that they better understand the circuit operation under constraints.
Unfortunately this generalization capability is mostly missing in
prior methods. Automated analog design has drawn increasing attention
to take human experts out of the optimization loop. Traditionally,
population based methods (e.g., genetic algorithm \cite{Cohen2015GeneticAS}
and particle swarm optimization \cite{Prajapati2015TwoSC}), and Bayesian
optimization \cite{Lyu2018AnEB,Lyu2018BatchBO} have been used. However,
the former suffers from low sample efficiency and lacks reproducibility.
The latter is sample efficient, but scales poorly in high dimensional
problems.

In this work, we introduce a novel reinforcement learning (RL) \cite{10.5555/3312046}
based optimization framework, DynaOpt, which not only learns the general
structure of solution space but also ensures high sample efficiency
based on a Dyna-style algorithm \cite{Sutton1991DynaAI}. The contributions
of this paper are as follows: First, the policy is trained through
a noise source and learns a whole distribution of feasible solutions.
Second, the reward is modeled using neural networks, which allows
us to leverage model-based RL. Third, the post-layout circuit is optimized
based on model-based RL, together with transfer learning from the
schematic based design, and we achieves 300$\times$ higher sample
efficiency than model-free approach. Finally, DynaOpt is implemented
by taking advantage of both the model-free and model-baed methods
to maximize the learning process from scratch. Even though we focus
on analog circuit design, the methodology is expected to be readily
applicable to a broad range of simulation based optimization problems
in engineering.

\section{Background}

\label{background}

\subsection{Circuit optimization problem}

The objective of analog circuit design is to search over optimal circuit
parameters (e.g., width and length of transistors, resistance and
capacitance) that meet design constraints imposed on performance metrics
(e.g., power, noise and voltage gain) given a circuit topology \cite{allen2002cmos}.
The performance metrics are measured by running a circuit simulator
with input of candidate parameters, and this process repeats until
all the required inequality constraints, which are application specific,
are satisfied (e.g., voltage gain > 100 V/V and power consumption
< 1mW).

The circuit optimization can be formulated as a RL problem. Generation
of circuit parameters corresponds to the action, and the reward is
determined by an estimate of circuit performance. The agent iterates
sequential decision process by interacting with the environment (i.e.,
circuit simulation), and improves the policy to maximize the expected
reward. We define the reward $R$ as a weighted sum of normalized
performance metrics under measurement.

\begin{equation}
R=\underset{i}{\sum}w_{i}r_{i}=\underset{i}{\sum}w_{i}\left[\min\left(\frac{m_{i}-m_{i}^{LOW}}{m_{i}+m_{i}^{LOW}},0\right)+\min\left(\frac{m_{i}^{UP}-m_{i}}{m_{i}^{UP}+m_{i}},0\right)\right]\label{eq:1}
\end{equation}

where the measured metric $m_{i}$ should be either larger than the
lower bound $m_{i}^{LOW}$ or smaller than the upper bound $m_{i}^{UP}$,
and $w_{i}$ denotes the weighting factor which is one in this work.
Each normalized metric $r_{i}$ is set to be within a range of $\left[-1,0\right]$,
which also helps regression of the reward model discussed in Section
\ref{dynaOpt}. The reward starts from the negative and reaches zero
once all hard constraints are satisfied. The other minimization or
maximization objectives can be also included in Equation \ref{eq:1}
similarly without clipping values to be negative.

\subsection{Related works}

Recent rapid progress in machine learning has accelerated analog design
automation with various learning based methods. For example, \cite{Hakhamaneshi2019BagNetBA}
combines evolutionary algorithms with a neural network discriminator
to boost the sample efficiency. Recent works \cite{Wang2018LearningTD,Settaluri2020AutoCktDR}
frame the optimization problem as a Markov decision process (MDP)
and apply the RL method to learn the optimal policy. RL is also used
together with graph convolutional neural networks to learn about the
circuit topology representation~\cite{Wang2020GCNRLCD}. However,
the aforementioned RL based approaches all rely on model-free algorithms
which require to run the circuit simulation for every training sample
to evaluate the reward, and thus the sample efficiency inevitably
degrades as the circuit complexity increases. In addition, \cite{Settaluri2020AutoCktDR}
attempts to learn about the design space through a sparse subsampling
technique (i.e., randomizing target specifications in a certain range),
but it may not guarantee convergence to generalized solutions.

\section{DynaOpt: reinforcement learning based optimization}

\label{dynaOpt}

In this section, we present a RL based optimization framework which
can tackle both the sample efficiency and the solution generalization.
The policy generator and the reward model are discussed in detail
as important building blocks in the proposed architecture, and it
is followed by a description of Dyna-style leaning based algorithm.

\subsection{Policy generator}

In this paper, we try to address the circuit optimization problem
with RL. Even though the problem has no time component, RL can still
be promising. For instance, RL has recently been used for the problem
of network architecture search (NAS) with good results \cite{Zoph2017NeuralAS}.
Our initial approach was to mirror their method: we used the recurrent
neural network (RNN) controller where each RNN cell generates one
circuit parameter at every time step, and it is then fed as input
to the next cell at the next time step. However, our experiments indicated
that the cell connections over time steps do not help for the problem
(see more details in Supplementary Material).

\begin{figure}
\centering{}\includegraphics[viewport=0bp 0bp 390.375bp 228bp,scale=0.45]{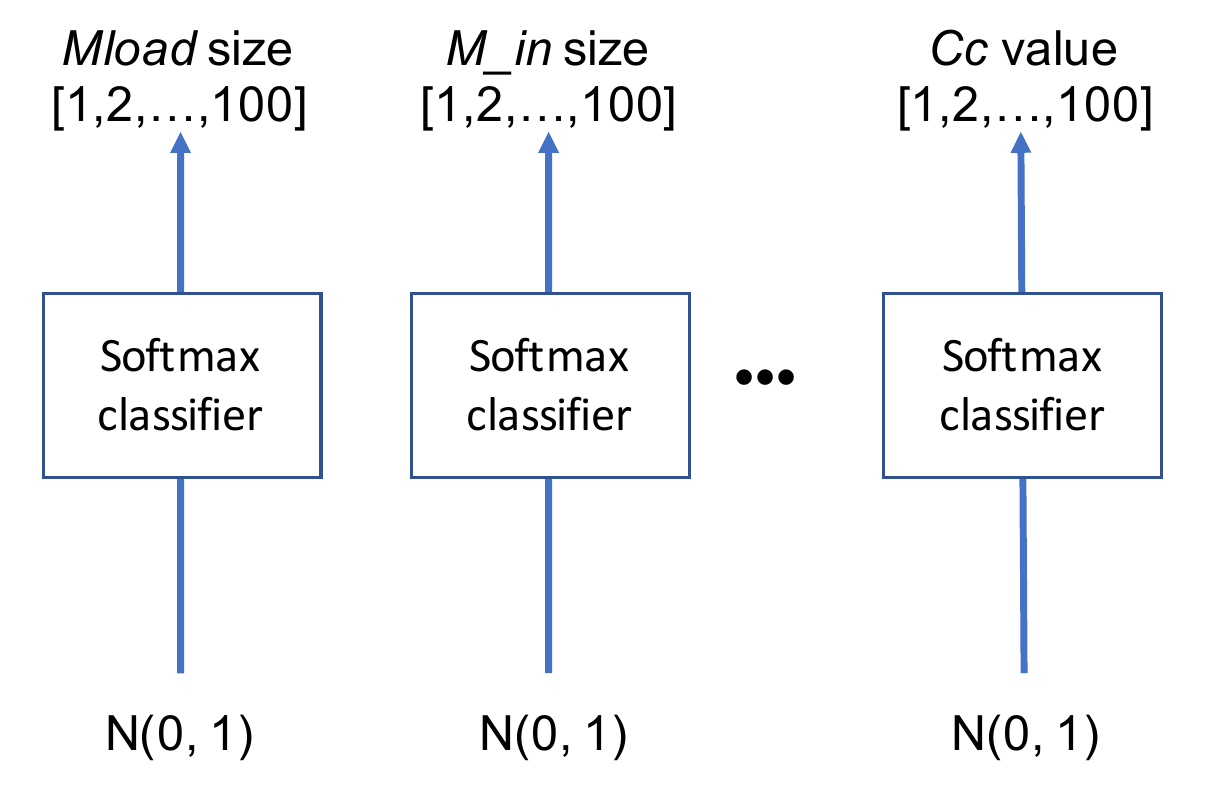}\includegraphics[scale=0.42]{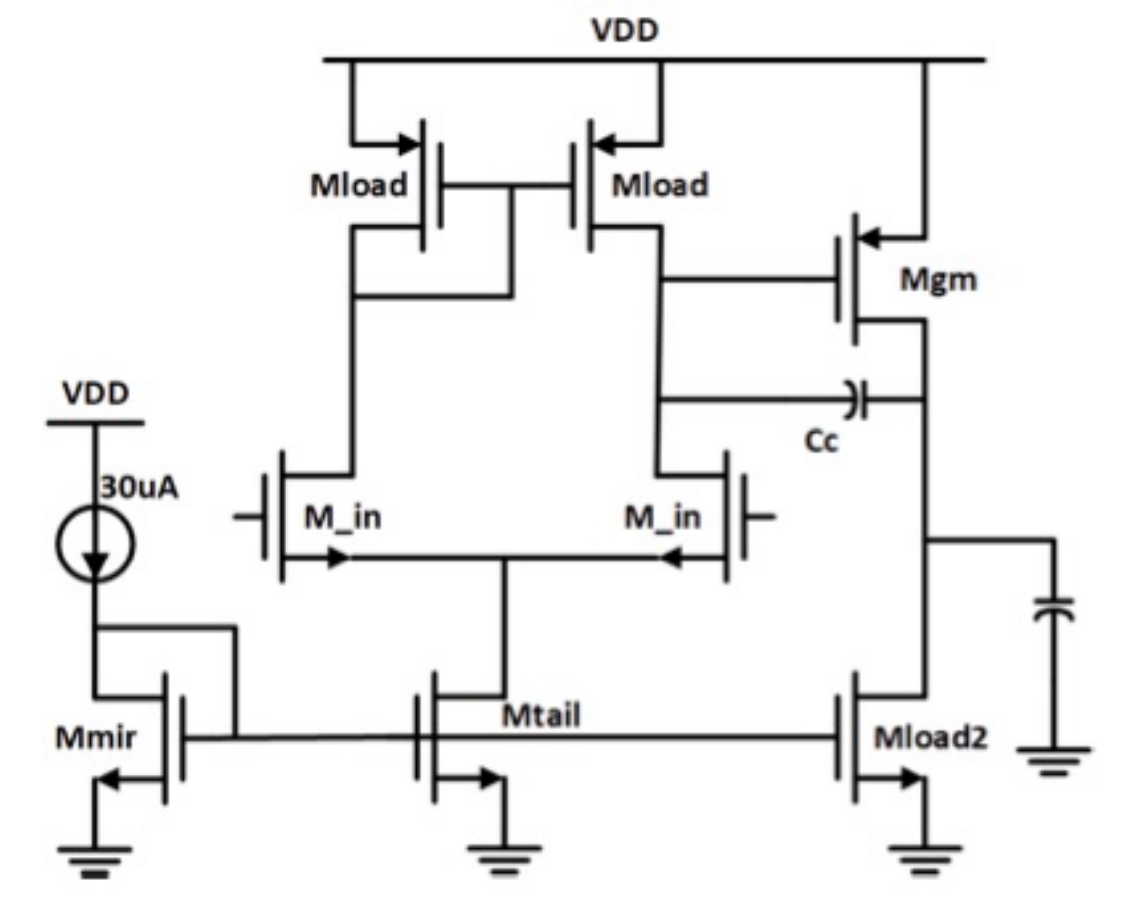}\caption{Left: Policy generator with input Gaussian noise, Right: Schematic
of two-stage operational amplifier \cite{Settaluri2020AutoCktDR}.}
\label{fig_policy_network}
\end{figure}

Therefore, we take a much simpler approach as illustrated in Figure
\ref{fig_policy_network} (left): each parameter is represented by
its own distribution. Together these form a policy that is a product
distribution: $\pi(\boldsymbol{a}|\boldsymbol{z})=\pi_{1}(a_{1}|z_{1})\cdot\ldots\cdot\pi_{T}(a_{T}|z_{T})$,
where $\boldsymbol{a}=\left\langle a_{1},\dots,a_{T}\right\rangle $
is a specification of all $T$ parameters. To allow flexible distributions
for each parameter, we draw inspiration from generative adversarial
networks (GANs) \cite{Goodfellow2014GenerativeAN}and use a feedforward
network which is fed by a Gaussian noise input such that it can ultimately
learn a distribution of optimal actions. For implementation, this
policy generator is trained using the REINFORCE policy gradient algorithm
\cite{Williams1992SimpleSG} with baseline, and the entropy bonus
is added to encourage exploration \cite{10.5555/3279266}.

\subsection{Reward model}

The reward evaluation with simulation is an expensive process in analog
circuit design, and here we investigate if it can be replaced by a
reward model built with learnable neural networks. The reward model
is simply a regression model predicting a scalar valued reward as
a function of actions. Alternately, it can be implemented to output
each performance metric separately, from which the reward is calculated
in the final stage. The neural network is designed to have different
heads and one shared backbone like a multi-task regressor. It can
serve to be better explainable to designers, but costs slightly increased
complexity in the neural network. Furthermore, since the environment
is fully represented by the reward model only, model-based RL can
be directly applied to the problem without incorporating any other
models. In this regard, the post-layout deign can greatly benefit
from model-baed RL, as discussed in detail below.

\paragraph{Transfer learning}

Design lifecycle of analog circuits generally involves two circuit
optimization tasks. While the first preliminary design considers the
schematic only, the post-layout design requires several costly EDA
tools to take into account the layout effect, and it poses a major
bottleneck. The overall design efficiency can be significantly improved
by transferring knowledge between the two, based on the model-based
approach. More specifically, it starts with optimizing the first design
with model-free RL. The training data of action-reward pairs are reused
to build up the reward model for the first design. Then the reward
model is tweaked for the post-layout design with a minimal number
of simulation based training samples. Finally the post-layout design
is optimized based on full model-based RL.

\subsection{Dyna-style optimization}

In case that prior knowledge related to the problem is unavailable,
model-based optimization can be implemented in a different way such
that the policy generator and the reward model are improved alternately
from scratch. At each training cycle, a small batch of action-reward
pairs sampled from the policy and evaluated with simulation, i.e.,
real experience, is accumulated to the sample buffer which is in turn
used to train the reward model. It is followed by the policy improvement
with this updated reward model. This iteration continues until the
optimal policy is reached. Moreover, since real experience is sampled
from the on-policy distribution, the reward model can be effectively
updated with much less training samples by focusing on the interested
action space defined by the current policy \cite{10.5555/3312046}.
Due to this advantage, the algorithm can be implemented in several
different ways such that, for instance, the reward model is trained
without using the off-policy sample buffer.

To further enhance the sample efficiency, a unified Dyna-style optimization
scheme is proposed by intermixing both the model-free and model-based
RL methods \cite{Sutton1991DynaAI}. The algorithm needs one modification
from the model-based one such that real experience is used to improve
the policy by model-free RL as well as the reward model. The details
are described in Algorithm \ref{algorithm_DynaOpt}.

\begin{algorithm}[h]
\caption{Proposed DynOpt method.}
\label{algorithm_DynaOpt}

Initialize policy generator $\pi_{\theta}\left(\left.\boldsymbol{a}\right|\boldsymbol{z}\right)$,
reward model $\rho_{\phi}\left(\boldsymbol{a}\right)$ and sample
buffer $B$

\textbf{For} number of training cycles \textbf{do}

.\qquad{}\textbf{For} $i=1:N_{direct}$ \textbf{do}\qquad{}// Loop
for model-free RL

.\qquad{}.\qquad{}Sample $\boldsymbol{z}\sim\mathcal{N}\left(\boldsymbol{0},\boldsymbol{1}\right)$

.\qquad{}.\qquad{}Sample a vector of $T$ actions $\boldsymbol{a}\sim\pi_{\theta}\left(\left.\cdot\right|\boldsymbol{z}\right)$\qquad{}//
$T$: number of circuit parameters

.\qquad{}.\qquad{}Estimate reward $R$ \emph{using circuit simulation
with $\boldsymbol{a}$}

.\qquad{}.\qquad{}Update policy parameters $\theta$ based on REINFORCE
rule

.\qquad{}.\qquad{}Store action-reward pair $\left(\boldsymbol{a},R\right)$
in $B$

.\qquad{}\textbf{End}

.\qquad{}Update reward model parameters $\phi$ by regression on
$B$

.\qquad{}\textbf{For} $i=1:N_{model}$ \textbf{do}\qquad{}// Loop
for model-based RL

.\qquad{}.\qquad{}Sample $\boldsymbol{z}\sim\mathcal{N}\left(\boldsymbol{0},\boldsymbol{1}\right)$

.\qquad{}.\qquad{}Sample $\boldsymbol{a}\sim\pi_{\theta}\left(\left.\cdot\right|\boldsymbol{z}\right)$

.\qquad{}.\qquad{}Estimate $R$ \emph{using reward model $\rho_{\phi}\left(\boldsymbol{a}\right)$}

.\qquad{}.\qquad{}Update $\theta$ based on REINFORCE rule

.\qquad{}\textbf{End}

\textbf{End}
\end{algorithm}

\section{Experiments}

\label{experiments}

An experiment is carried out on the same design problem of a two-stage
operational amplifier as found in \cite{Settaluri2020AutoCktDR}.
The circuit schematic is shown in Figure \ref{fig_policy_network}
(right). The problem has 7 circuit parameters (6 transistor sizes,
1 capacitance) to optimize and 4 design constraints (voltage gain
> 200 V/V, unity gain bandwidth > 1 MHz, phase margin > 60 degree,
bias current < 10 mA) to satisfy. Each discretized circuit parameter
can be selected from 100 possible values, and so the action space
size is $10^{14}$ in total. The circuit is evaluated using 45nm BSIM
model with NGSPICE simulator \cite{GithubAutoCkt}.

\subsection{Analysis of policy generator}

\begin{figure}
\centering{}\includegraphics[scale=0.24]{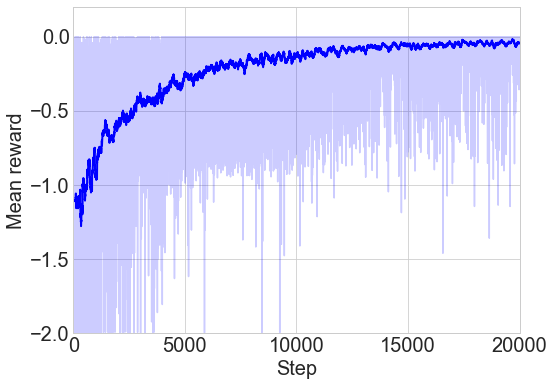}\includegraphics[scale=0.24]{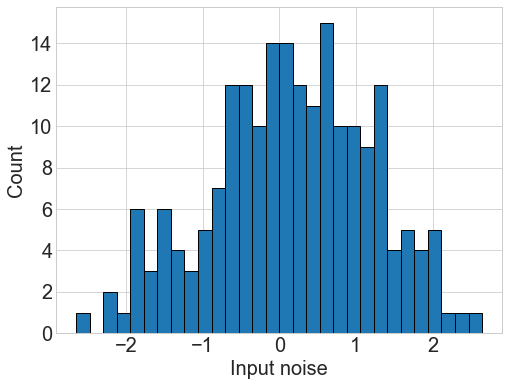}\includegraphics[scale=0.24]{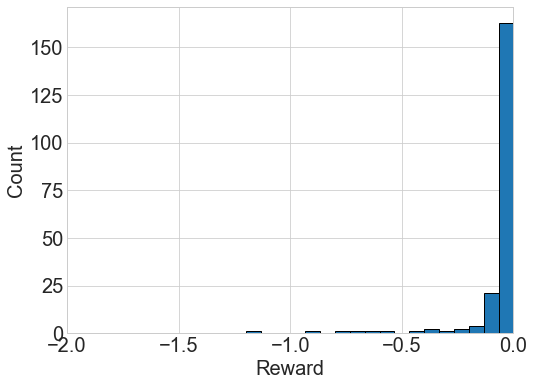}\caption{Left: Mean reward curve in policy training, Center: Distribution of
Gaussian noise input to policy generator, Right: Simulation based
evaluated reward distribution of generated actions from 200 noise
samples.}
\label{fig_policy_generator}
\end{figure}

We train the policy generator in Figure \ref{fig_policy_network}
(left) based on the policy gradient method, using random noise input
from the normal distribution $\mathcal{N}\left(0,1\right)$ for generalization.
Figure \ref{fig_policy_generator} (left) shows a noisy reward trajectory
to learn about the action space from the noise input, and the mean
reward converges to zero around 20,000 steps\footnote{Because a minibatch size of one is used throughout this work, the
training step in model-free algorithms is equal to the number of simulation.} as the policy improves. For verification, this trained policy generator
is evaluated with 200 simulation based samples given by the same noise
source as found in Figure \ref{fig_policy_generator} (center). Figure
\ref{fig_policy_generator} (right) shows the reward distribution
of the corresponding actions generated by the trained policy and evaluated
by simulation. The generated actions are distributed mostly near a
reward of zero, and eventually the policy generator can learn to map
the input noise to a distribution of optimal actions solving the problem.

\subsection{Transfer learning for post-layout design}

\begin{figure}[b]
\begin{centering}
\includegraphics[viewport=0bp 0bp 634bp 386bp,scale=0.235]{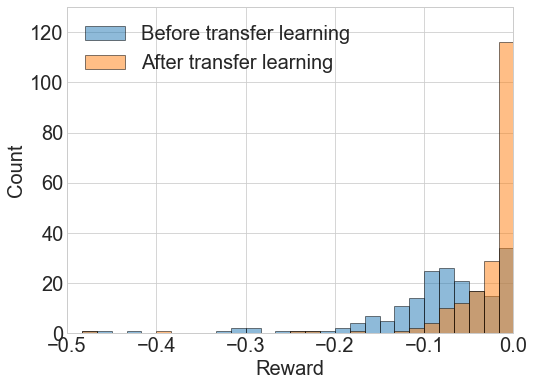}\includegraphics[scale=0.235]{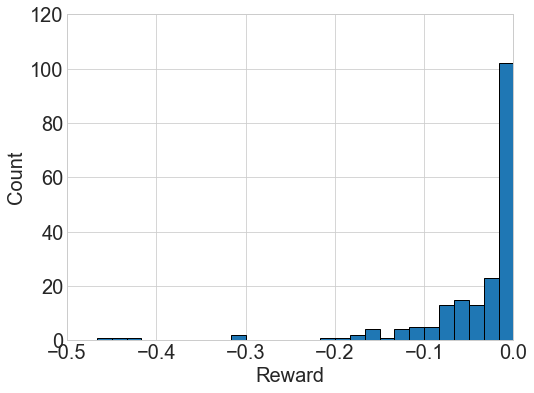}
\par\end{centering}
\caption{Post-layout circuit optimization: simulation based evaluated reward
distribution obtained with model-based RL optimization and transfer
learning with 100 training samples (left) and model-free RL optimization
results with 30,000 training samples (right).}
\label{fig_transfer_learning}
\end{figure}

In the post-layout design, we simply assume that the layout effect
introduces additional parasitic capacitance of 200 pF between all
nodes to the circuit in Figure \ref{fig_policy_network} (right).
This value is likely to be much larger than actual ones like some
worse-case scenario, and accurate assessment of the layout parasitics
may not be critical here for proof-of-principle experiment. Figures
\ref{fig_transfer_learning} (left) shows the model-based optimization
result. Initially the policy trained in the first schematic-only design
as described above, generates low-quality actions unsuitable for this
post-layout circuit. Only 100 simulation based training samples are
used to update the reward model by transfer learning. The reward model
is configured to have 3 hidden layers of size 16 with ReLU activation
except the output layer. The policy is then updated from the pretrained
policy and optimized for the post-layout circuit, based on model-based
RL without involving simulation at all. The reward distribution is
obtained by evaluating the trained policy with 200 input noise samples
as before. For comparison, model-free optimization is also carried
out by learning from scratch with 30,000 simulations, and the result
is shown in Figure \ref{fig_transfer_learning} (right). The model-based
approach achieves comparable performance, with 300$\times$ improved
sample efficiency.

\subsection{DynaOpt based circuit optimization}

\begin{figure}
\begin{centering}
\includegraphics[scale=0.33]{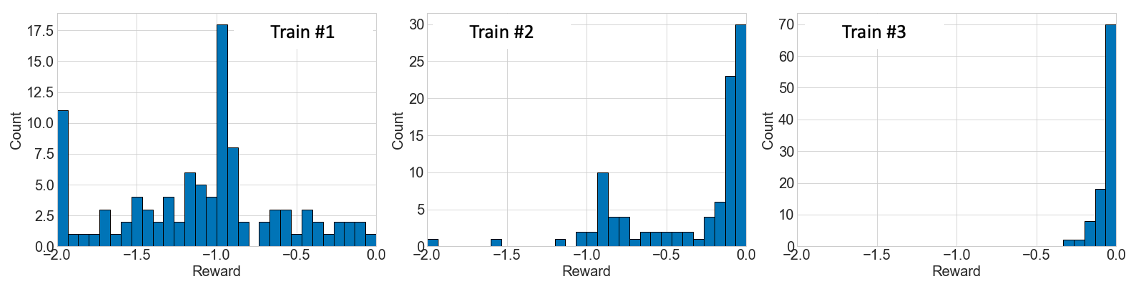}
\par\end{centering}
\begin{centering}
\includegraphics[scale=0.33]{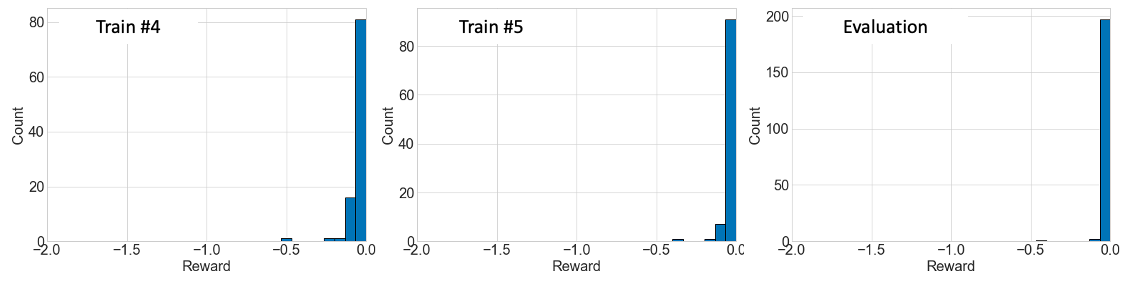}
\par\end{centering}
\caption{DynaOpt based optimization of two-stage operational amplifier: evolution
of simulation based reward distributions measured at each training
cycle (100 samples) and final evaluation (200 samples).}
\label{fig_DynaOpt}
\end{figure}

DynaOpt is also applied to the problem in Figure \ref{fig_policy_network}
(right). The training is repeated over 5 cycles, and each cycle uses
100 simulations to update both the policy generator and the reward
model, based on Dyna-style RL, as described in Algorithm \ref{algorithm_DynaOpt}.
Figure \ref{fig_DynaOpt} shows how the reward distribution evolves
as the policy improves, when the agent starts learning without prior
knowledge. As compared to Figure \ref{fig_policy_generator} (right)
where the the model-free method is applied with 20,000 simulations,
DynaOpt yields comparable results only after about 3 training cycles
(300 simulations), and learns nearly optimal policy with merely 500
simulations.

\section{Conclusions}

\label{conclusions}

We propose DynaOpt for analog circuit design, which is a Dyna-style
RL based optimization framework. It is built by intermixing both the
model-free and model-based methods with two key components - the stochastic
policy generator and the reward model. The policy generator learns
to map the random noise input to the on-policy distribution of actions
which eventaully move toward solutions as the policy improves, and
the reward model allows to leverage sample efficient model-based RL
by replacing expensive simulation with it. By putting them together,
DynaOpt achieves both generalization capability and high sample efficiency,
which is difficult with prior methods. Application to the design of
a two-stage operational amplifier is demonstrated based on various
implementation of the methodology such as model-based learning with
knowledge transfer and Dyna-style learning, which all outperform the
model-free approach with promising results.

\begin{ack}
  \begin{minipage}{\linewidth}
  \begin{wrapfigure}{r}{0.26\columnwidth}
    \centering
    \vspace{-20pt}
    \includegraphics[scale=0.1]{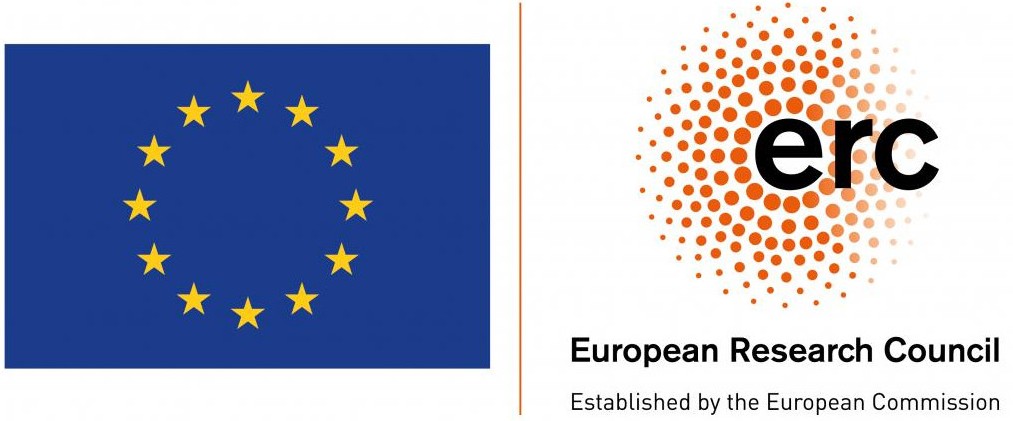}
  \end{wrapfigure}
  This project had received funding from the European Research Council (ERC) under the European Union's Horizon 2020 research and innovation programme (grant agreement No.~758824 \textemdash INFLUENCE).
  \end{minipage}
\end{ack}

\bibliographystyle{unsrt}
\bibliography{ml4eng_2020_DynaOpt}

\part*{\newpage}

\section{Supplementary Material}

\begin{figure}[ph]
\centering{}\includegraphics[scale=0.45]{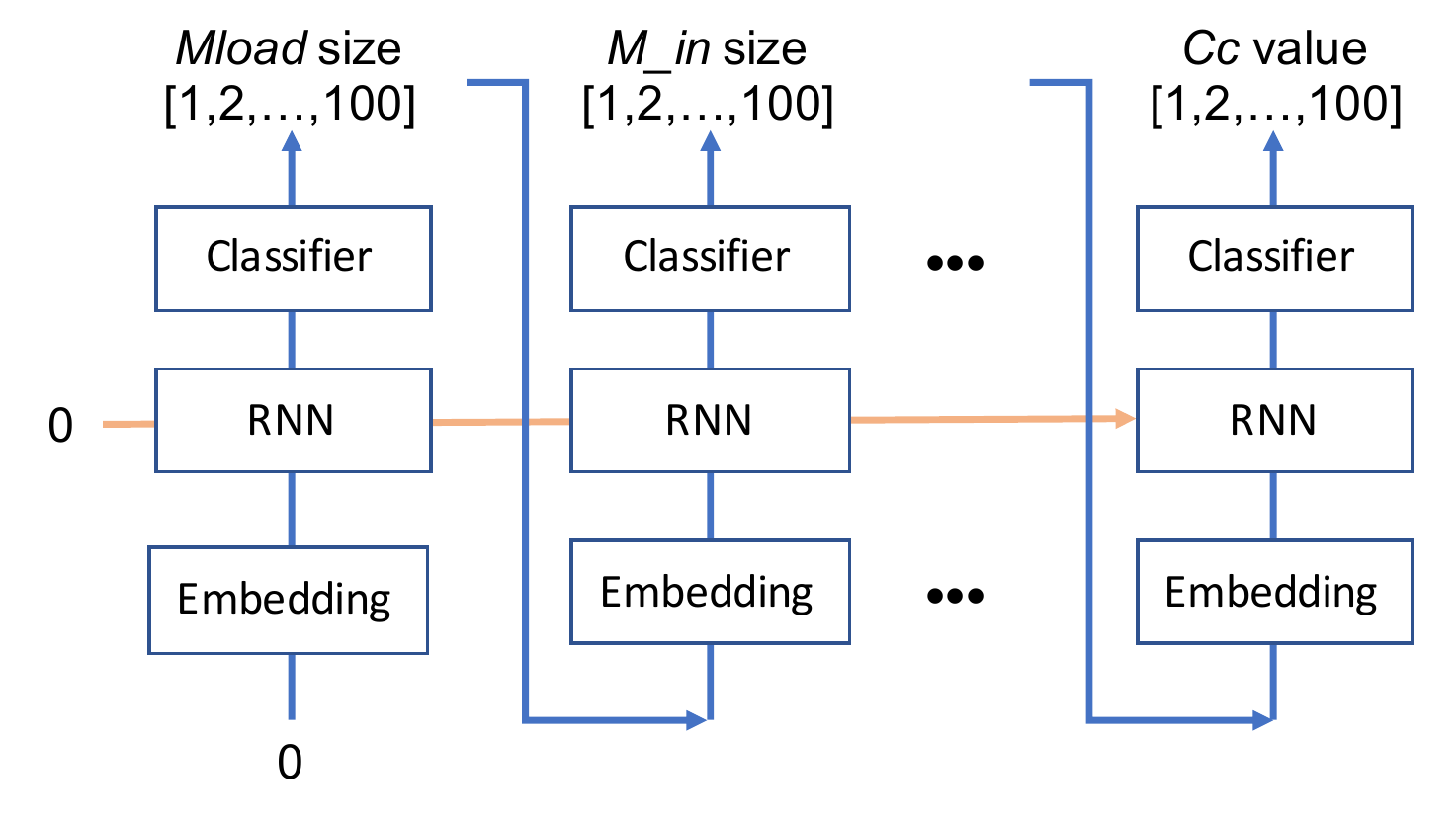}\caption{RNN controller for circuit optimization.}
\label{fig_supplement_1}
\end{figure}

\begin{figure}[ph]

\centering{}\includegraphics[viewport=0bp 0bp 649bp 386bp,scale=0.24]{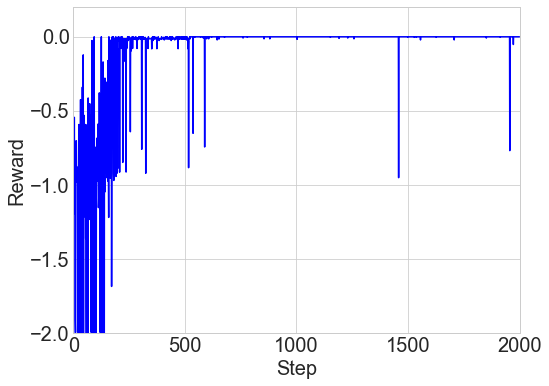}\includegraphics[scale=0.24]{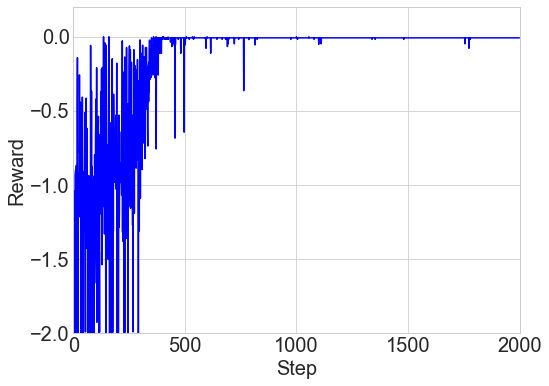}\caption{Left: Reward curve obtained by directly applying the RNN controller,
Right: Reward curve obtained by disconnecting cells in the controller.}
\label{fig_supplement_2}
\end{figure}

Due to similarity to the NAS problem, the RL based approach in \cite{Zoph2017NeuralAS}
is directly applicable to the circuit optimization as long as the
action space is discretized. As shown in Figure \ref{fig_supplement_1},
the RNN controller is used as the policy network to sequentially sample
a list of actions, i.e., circuit parameters, based on the current
policy. The action generated in each RNN cell is fed to the next cell
in an autoregressive way. After collecting the final action, the agent
receives the corresponding reward by measuring the circuit performance
with simulation. We first apply the RNN controller to the circuit
problem in Figure \ref{fig_policy_network} (right), and Figure \ref{fig_supplement_2}
(left) shows the reward over training steps. The problem is solved,
i.e., approaching a reward of zero, in < 500 steps, with producing
one random combination of optimal parameters. Different orders of
parameters assigned to RNN cells are tested, and they produce all
similar learning curves without any preferred orders observed.

However, the controller still works similarly even when the cells
are disconnected by deactivating the RNN (i.e., setting zero to all
inputs and hidden states) and cutting off the autoregressive connection,
as shown in Figure \ref{fig_supplement_2} (right). Each cell works
as an independent learner like treating the circuit parameters as
independent variables, and thus the policy network for the circuit
optimization can be greatly simplified to the scheme presented in
Figure \ref{fig_policy_network} (left).

\end{document}